\documentclass[a4paper]{article}
\usepackage{onecolceurws}

\usepackage{booktabs} %

\usepackage[T1]{fontenc}
\usepackage[utf8]{inputenc}
\usepackage{lmodern}
\usepackage{wrapfig}
\usepackage{subfig}
\usepackage[ruled,vlined,linesnumbered]{algorithm2e}
\usepackage{fancyvrb}
\usepackage{amsmath}
\usepackage{amsfonts}
\usepackage{graphicx}

\usepackage{xcolor}

\usepackage{footnote}
\makesavenoteenv{table}
\makesavenoteenv{tabular}
\makesavenoteenv{table*}
\makesavenoteenv{tabular*}

\usepackage{tikz}
\usetikzlibrary{shapes}

\usepackage{forest}

\usepackage{dsfont}

\usepackage{microtype}
\VerbatimFootnotes
\usepackage{todonotes}
\usepackage{soul}
\newcounter{mcomment}

\newcounter{fpcomment}

	 \title{Towards Formula Translation using\\ Recursive Neural Networks}

	\author{Felix Petersen \and Moritz Schubotz \and Bela Gipp \\
	}
	\institution{
	Department of Computer and Information Science\\University of Konstanz, Germany\\
	first.last@uni-konstanz.de}%

\begin{document}

	\maketitle

	\begin{abstract}
		
		While it has become common to perform automated translations on natural language, performing translations between different representations of mathematical formulae has thus far not been possible.
		We implemented the first translator for mathematical formulae based on recursive neural networks.
		We chose recursive neural networks because mathematical formulae inherently include a structural encoding.
		In our implementation, we developed new techniques and topologies for recursive tree-to-tree neural networks based on multi-variate multi-valued Long Short-Term Memory cells.
		We propose a novel approach for mini-batch training that utilizes clustering and tree traversal.
		We evaluate our translator and analyze the behavior of our proposed topologies and techniques based on a translation from \textit{generic} \LaTeX{} to the \textit{semantic} \LaTeX{} notation.
		We use the \textit{semantic} \LaTeX{} notation from the Digital Library for Mathematical Formulae and the Digital Repository for Mathematical Formulae at the National Institute for Standards and Technology.
		We find that a simple heuristics-based clustering algorithm outperforms the conventional clustering algorithms on the task of clustering binary trees of mathematical formulae with respect to their topology.
		Furthermore, we find a mask for the loss function, which can prevent the neural network from finding a local minimum of the loss function.
		Given our preliminary results, a complete translation from formula to formula is not yet possible.
		However, we achieved a prediction accuracy of $47.05\%$ for predicting symbols at the correct position and an accuracy of $92.3\%$ when ignoring the predicted position.
		Concluding, our work advances the field of recursive neural networks by improving the training speed and quality of training.
		In the future, we will work towards a complete translation allowing a machine-interpretation of \LaTeX{} formulae.

	\end{abstract}

\clearpage
	\section{Introduction}\label{sec.intro}

	Most mathematical formulae are denoted in presentation language (PL)~\cite{SchubotzGSMCG18} for the purpose of displaying them.
	However, PL does not allow machine-interpretation of the formula, i.e. any semantic information contained in a given formulae cannot be understood by Computer Algebra Systems (CAS).
	One reason for this limitation is the ambiguity of PL.
	For example, $\bar{x}$ (\verb@\bar{x}@) may either be the mean of $x$, the complex conjugate of $x$, or it may take on an array of other meanings.
	On the other hand, for humans, the machine-readable unambiguous notations of content language (CL) are often more laborious to produce, read, and remember.
	Therefore, a reliable machine translation between PL and CL is a crucial step towards the automatized machine-interpretation of mathematical notation that is found in academic and technical documents.

	The National Institute for Standards and Technology (NIST), the Digital Library for Mathematical Formulae (DLMF), and the Digital Repository for Mathematical Formulae (DRMF) developed a set of \LaTeX{} macros that allow an unambiguous mathematical notation within \LaTeX{} \cite{Miller2003}.
	We refer to this notation as \textit{semantic} \LaTeX{} and refer to the standard \LaTeX{} maths notation as \textit{generic} \LaTeX{}.
	It is possible to translate from \textit{semantic} \LaTeX{} to \textit{generic} \LaTeX{} using \LaTeX{}ML.
	However, the reverse translation from \textit{generic} to \textit{semantic} \LaTeX{} is not solved yet.
	To achieve this reverse translation the disambiguation of mathematical formulae is mandatory.
	
	Thus, we experimented with a standalone neural disambiguation that classifies which \textit{semantic} \LaTeX{} macro is most likely to be correct.
	To use this approach, we would require a rule-based translator that utilizes the neural network only in ambiguous cases.

	Rule-based translators are not readily transferable to other mathematical languages.
	Furthermore, they require a very complex program that depends on the input and output language and are also hard to maintain if the languages are extended or a new language should be added.

	Therefore, we use machine learning methods to build a full translator as announced in~\cite{Cohl15}.
	How well implicit mathematical disambiguation within a machine learning-based translator performs remains is subject of future research.
	We expect that our new approach outperforms existing approaches when it comes to the implicit mathematical disambiguation, since simple neural networks already yielded significantly better results than traditional approaches on similar tasks as discussed below.
	Recently, recurrent neural networks with sequence-to-sequence models yielded state-of-the-art results on natural language translation tasks \cite{DBLP:journals/corr/Lipton15}. 
	In natural language machine translation, even an inaccurate paraphrasing translation is sufficient to satisfy a user's most basic information needs.
	However, in contrast, inaccurate translations in mathematical machine translation are most likely useless and misleading for the typical mathematical information needs.
	Thus, in contrast to natural language, for mathematical formulae a precise translation in mandatory.
	As sequential recurrent neural networks don't utilize the tree-like structure of input and output, they lack precision in the domain of structure.

	An alternative to recurrent neural networks is recursive neural networks, which are not linear but have a tree-like structure.
	By building a recursive neural network, the neural network explicitly receives information on the location of the respective words and symbols.
	As soon as this information is relevant for the translation, a recursive neural network can yield better results than a recurrent neural network.
	Further, recursive neural networks have a lower depth than recurrent neural networks.
	Thus, the problem of exploding or vanishing gradients is reduced.

	In most cases, no structural information about natural language is given.
	Therefore, it is required to deduce a structure for these sentences.
	Finding such a structure is ambiguous and error-prone.
	Mathematical formulae often allow a straightforward derivation of the semantic structure or are already given in a structured format.
	Thus, we believe that utilizing recursive neural networks is a promising approach for the automatic translation of mathematical formulae.

	In summary, our research contributes towards:
	\begin{itemize}
		\item the \textbf{disambiguation} of mathematical formulae, and 
		\item the \textbf{neural machine translation} of mathematical formulae.
	\end{itemize}
	Furthermore, our research yields the following contributions for recursive neural networks:
	\begin{itemize}
		\item \textbf{faster training} by using a semi-batched computation, and
		\item a \textbf{local minima avoiding} loss function for decoders.
	\end{itemize}
	The contributions for recursive neural networks may also be applicable to other machine-learning tasks, including sentiment analysis.

	We structure the presentation of our contributions as follows.
	Section \ref{sec.related_work} presents an overview of prior research.
	Section \ref{sec.method} states the problem of translation and presents our neural network approach.
	Section \ref{sec.data} presents the applied preprocessing steps for the formulae.
	Section \ref{sec.results} presents our evaluation techniques and current results.
	Section \ref{sec.disconcl} discusses our results and conclusions.
	Finally, avenues for future work are presented in section \ref{sec.outlook}.

	\section{Related Work}\label{sec.related_work}

	To the best of our knowledge, no research towards machine learning translation for mathematical formulae between different representations has been performed thus far.
	Our research towards machine translation of mathematical formulae will enable future work on various topics from automated checking of mathematical formulae to plagiarism detection.
	There have already been rule-based attempts to translate between mathematical languages.
	There exists a rule-based translator from \textit{semantic} \LaTeX{} to CASs (e.g.,~Maple) \cite{Cohl17}.
	Thus, by translating from \textit{generic} to \textit{semantic} \LaTeX{}, we also obtain a notation that is readable by CASs.
	The \textit{semantic} \LaTeX{} to Maple translator achieved an accuracy of $53.59\%$ on correctly translating $4\,165$ test equations from the DLMF.
	The accuracy of the \textit{semantic} \LaTeX{} to CAS translator is relatively low because of the high complexity of these test equations, and because most of the functions which are represented by a DLMF/DRMF \LaTeX{} macro are not defined in Maple \cite{Cohl17}.
	Our approach may act as a substitute for this translator in cases where the aspired CAS is not yet implemented.
	The \textit{semantic} \LaTeX{} to CAS translator may also be used as a reference for our approach.

	State-of-the-art for language translation uses recurrent neural networks employing, e.g., Long Short-Term Memory (LSTM) cells in sequence-to-sequence networks \cite{tf-seq2seq}.

	In recent time, also recursive neural networks have been used for the translation of structured data in tree-to-tree networks.
	Chen et al.~used tree-to-tree networks for program translation \cite{DBLP:journals/corr/abs-1802-03691}.
	Recursive neural networks have mainly been used for different machine learning tasks, such as sentiment analysis of the Stanford sentiment tree bank \cite{DBLP:journals/corr/TaiSM15}, or classification of collimated sprays of energetic hadrons in quantum chromodynamics \cite{Louppe:2017ipp}.
	A majority of tasks does not require a random sized and structured output but rather a classification.
	Therefore, these neural networks only need a recursive encoder.
	Thus, only a minority of research has been performed on recursive decoders.
	Local minima of loss functions often cause neural networks to stop learning or even lead them to predict only one constant value.
	Thus, we propose a local minimum avoiding loss function for the training of recursive decoders.
	Tai et al.~\cite{DBLP:journals/corr/TaiSM15} presented tree-structured LSTM networks.
	LSTMs are building blocks of neural networks that are composed of a cell, an input gate, an output gate and a forget gate.
	Their purpose is to remember values over a long time and avoid the problem of exploding gradients utilizing their forget gate.
	We extend this approach using multi-variate multi-valued LSTMs in order to build recursive LSTM networks with a recursive encoder, as well as a recursive decoder.
	
	This means that we use a neural network whose encoder and decoder have the same topology as the binary parse tree of the input formula respectively output formula.
	Therefore, the structure of the input defines the topology of the neural network.
	Thus, in general, for the training of recursive neural networks, a batched computation is not directly possible.
	A reason for that is that trees with different topologies usually cannot share one neural network \cite{DBLP:journals/corr/BowmanGRGMP16}.
	Alternatively, the neural network could be made big enough to capture the whole dataset, which would result in neural networks with tens or hundreds of thousands of LSTM-units.
	Since one of the main issues of neural networks is long training times, a speedup of the training is essential.
	Recursive neural networks especially suffer from long training times since a batched computation is not directly possible \cite{DBLP:journals/corr/BowmanGRGMP16}.
	Thus, we propose a new technique that enables semi-batched computation for recursive neural networks utilizing clustering and tree traversal.

	\section{Method}\label{sec.method}
	
	In this section, we will present the problem of translation of trees and give an overview of our approach.
	Further, we will present our method for a standalone disambiguation and comment on our implementation.
	
	\subsection{Problem Statement}
	Let $\textbf{x}\in \textbf{X}$ be the binary parse tree of a \textit{generic} \LaTeX{} formula. Further, let $\textbf{y}\in \textbf{Y}$ be the respective binary parse tree of the \textit{semantic} \LaTeX{} translation of $\textbf{x}$.
	In these binary trees, only the leaves are labeled with positive integers corresponding to the words and symbols of the formula, all other nodes are labeled as zero.
	
	Further, let ${\textbf{x}}_a$ for $a$, being a binary tree with $a \supset \textbf{x}$, be a tree with the topology of $a$ and the values of $\textbf{x}$ padded with the tag $\textbf{y}_\mathrm{end}$ with respect to the topology of $a$.
	Here, ``$a \supset b$'' means that for each node in $b$ there is a node in $a$ at the same position in the tree.
	The padding is performed such that a position is tagged with $\textbf{y}_\mathrm{end}$ iff there is no node at the respective position in $\textbf{x}$.
	
	We want to find a mapping $f: \textbf{X} \to \textbf{Y}, \textbf{x} \mapsto \textbf{y}$.
	
	This mapping has to be derivable with respect to the tree topologies, the definition of our recursive neural network, and the trained weight and bias matrices.
	
	Therefore, we create a neural network and generate a loss as a distance measure between the ground truth and the prediction.
	By minimizing the loss, we get an improved mapping.

	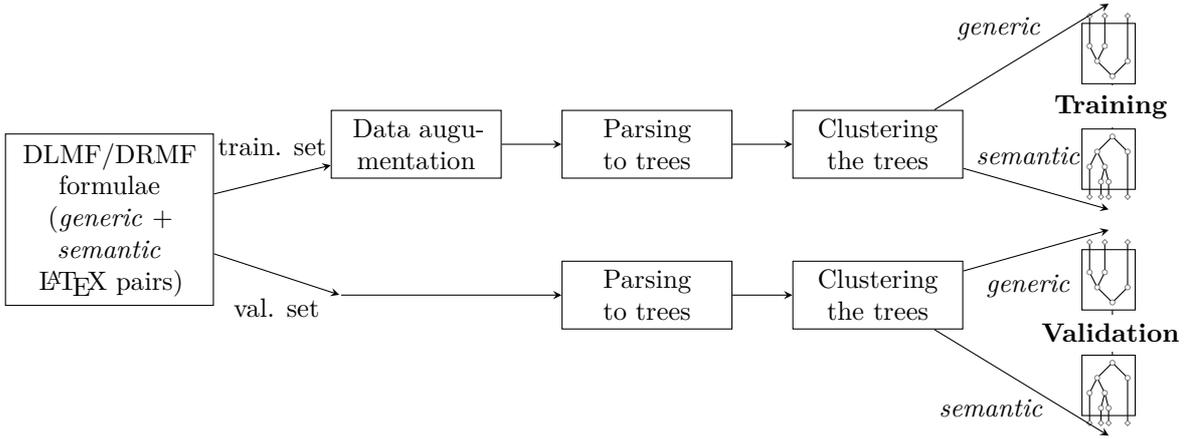
\begin{figure}
		\centering
		
		\usetikzlibrary{arrows}
		\def\layersep{1}
		\def\stepoffset{4}

		\begin{tikzpicture}[node distance=\layersep,scale=0.65]
		
		\tikzstyle{arrow} = [->,>=stealth]
		
		\node[rectangle, text centered, text width=2.5cm, draw=black](data) at (0,0) {DLMF/DRMF formulae (\textit{generic} + \textit{semantic} \LaTeX{} pairs)};
		
		\node[rectangle, text centered, text width=2cm, draw=black, right of=data, xshift=4cm, yshift=1cm](aug) {Data augumentation};
		\node[right of=data, xshift=3cm, yshift=-1cm, scale=0.1](no-aug) {};
		
		\draw [arrow] (data) -- node[anchor=south, yshift=0.2cm] {train. set} (aug);
		\draw [arrow] (data) -- node[anchor=north, yshift=-0.2cm] {val. set} (no-aug);
		
		\node[rectangle, text centered, text width=2cm, draw=black, right of=aug, xshift=3cm](parse1) {Parsing to trees};
		\node[rectangle, text centered, text width=2cm, draw=black, right of=no-aug, xshift=4cm](parse2) {Parsing to trees};
		
		\draw [arrow] (aug) -- (parse1);
		\draw [arrow] (no-aug) -- (parse2);
		
		\node[rectangle, text centered, text width=2cm, draw=black, right of=parse1, xshift=3cm](cluster1) {Clustering the trees};
		\node[rectangle, text centered, text width=2cm, draw=black, right of=parse2, xshift=3cm](cluster2) {Clustering the trees};
		
		\draw [arrow] (parse1) -- (cluster1);
		\draw [arrow] (parse2) -- (cluster2);

		\node[right of=cluster1, xshift=3cm, yshift=0.5cm](tree1) {
			\begin{tikzpicture}[node distance=\layersep,scale=0.2]
			\tikzstyle{neuron} = [circle, draw=black!50, line width=0.1mm, fill=white, scale=0.2];
			\tikzstyle{treenode} = [circle, draw=black!50, line width=0.1mm, fill=white, scale=0.2];
			\tikzstyle{particle} = [diamond, draw=black!50, line width=0.1mm, fill=white, scale=0.15];

			\node[treenode] (tree1-h1) at (0,0) {};
			\node[treenode] (tree1-h2) at (-1,-1) {};
			\node[treenode] (tree1-h3) at (1,-1) {};
			\path[black] (tree1-h2) edge (tree1-h1);
			\path[black] (tree1-h3) edge (tree1-h1);
			\node[treenode] (tree1-h4) at (-1.5,-2) {};
			\node[treenode] (tree1-h5) at (-0.5,-2) {};
			\path[black] (tree1-h4) edge (tree1-h2);
			\path[black] (tree1-h5) edge (tree1-h2);
			\node[treenode] (tree1-h6) at (-0.75,-3) {};
			\node[treenode] (tree1-h7) at (-0.25,-3) {};
			\path[black] (tree1-h6) edge (tree1-h5);
			\path[black] (tree1-h7) edge (tree1-h5);
			\draw (-2,-3.5) rectangle (1.5, 0.5);
			
			\node[particle] (t1-v1) at (-1.5,-4) {};
			\path[black] (t1-v1) edge (tree1-h4);
			\node[particle] (t1-v2) at (-0.75,-4) {};
			\path[black] (t1-v2) edge (tree1-h6);
			\node[particle] (t1-v3) at (-0.25,-4) {};
			\path[black] (t1-v3) edge (tree1-h7);
			\node[particle] (t1-v4) at (1,-4) {};
			\path[black] (t1-v4) edge (tree1-h3);
			
			\node[neuron] (gru1) at (0,1.5) {};
			\path[black] (0,0.5) edge (gru1);
			\node[treenode] (tree2-h1) at (0,\stepoffset-0) {};
			\node[treenode] (tree2-h2) at (-1,\stepoffset--1) {};
			\node[treenode] (tree2-h3) at (1,\stepoffset--1) {};
			\path[black] (tree2-h2) edge (tree2-h1);
			\path[black] (tree2-h3) edge (tree2-h1);
			\node[treenode] (tree2-h4) at (-1.5,\stepoffset--2) {};
			\node[treenode] (tree2-h5) at (-0.5,\stepoffset--2) {};
			\path[black] (tree2-h4) edge (tree2-h2);
			\path[black] (tree2-h5) edge (tree2-h2);
			\draw (-2,\stepoffset+3.5) rectangle (1.5, \stepoffset-0.5);
			
			\node[particle] (t2-v1) at (2.5-4,\stepoffset+4) {};
			\path[black] (t2-v1) edge (tree2-h4);
			\node[particle] (t2-v2) at (3.5-4,\stepoffset+4) {};
			\path[black] (t2-v2) edge (tree2-h5);
			\node[particle] (t2-v3) at (5-4,\stepoffset+4) {};
			\path[black] (t2-v3) edge (tree2-h3);

			\path[black] (0,\stepoffset-0.5) edge (gru1);
			\end{tikzpicture} 
		};
	
		\node[right of=cluster2, xshift=3cm, yshift=-0.5cm](tree2) {
			\begin{tikzpicture}[node distance=\layersep,scale=0.2]
			\tikzstyle{neuron} = [circle, draw=black!50, line width=0.1mm, fill=white, scale=0.2];
			\tikzstyle{treenode} = [circle, draw=black!50, line width=0.1mm, fill=white, scale=0.2];
			\tikzstyle{particle} = [diamond, draw=black!50, line width=0.1mm, fill=white, scale=0.15];

			\node[treenode] (tree1-h1) at (0,0) {};
			\node[treenode] (tree1-h2) at (-1,-1) {};
			\node[treenode] (tree1-h3) at (1,-1) {};
			\path[black] (tree1-h2) edge (tree1-h1);
			\path[black] (tree1-h3) edge (tree1-h1);
			\node[treenode] (tree1-h4) at (-1.5,-2) {};
			\node[treenode] (tree1-h5) at (-0.5,-2) {};
			\path[black] (tree1-h4) edge (tree1-h2);
			\path[black] (tree1-h5) edge (tree1-h2);
			\node[treenode] (tree1-h6) at (-0.75,-3) {};
			\node[treenode] (tree1-h7) at (-0.25,-3) {};
			\path[black] (tree1-h6) edge (tree1-h5);
			\path[black] (tree1-h7) edge (tree1-h5);
			\draw (-2,-3.5) rectangle (1.5, 0.5);
			
			\node[particle] (t1-v1) at (-1.5,-4) {};
			\path[black] (t1-v1) edge (tree1-h4);
			\node[particle] (t1-v2) at (-0.75,-4) {};
			\path[black] (t1-v2) edge (tree1-h6);
			\node[particle] (t1-v3) at (-0.25,-4) {};
			\path[black] (t1-v3) edge (tree1-h7);
			\node[particle] (t1-v4) at (1,-4) {};
			\path[black] (t1-v4) edge (tree1-h3);
			
			\node[neuron] (gru1) at (0,1.5) {};
			\path[black] (0,0.5) edge (gru1);
			\node[treenode] (tree2-h1) at (0,\stepoffset-0) {};
			\node[treenode] (tree2-h2) at (-1,\stepoffset--1) {};
			\node[treenode] (tree2-h3) at (1,\stepoffset--1) {};
			\path[black] (tree2-h2) edge (tree2-h1);
			\path[black] (tree2-h3) edge (tree2-h1);
			\node[treenode] (tree2-h4) at (-1.5,\stepoffset--2) {};
			\node[treenode] (tree2-h5) at (-0.5,\stepoffset--2) {};
			\path[black] (tree2-h4) edge (tree2-h2);
			\path[black] (tree2-h5) edge (tree2-h2);
			\draw (-2,\stepoffset+3.5) rectangle (1.5, \stepoffset-0.5);
			
			\node[particle] (t2-v1) at (2.5-4,\stepoffset+4) {};
			\path[black] (t2-v1) edge (tree2-h4);
			\node[particle] (t2-v2) at (3.5-4,\stepoffset+4) {};
			\path[black] (t2-v2) edge (tree2-h5);
			\node[particle] (t2-v3) at (5-4,\stepoffset+4) {};
			\path[black] (t2-v3) edge (tree2-h3);

			\path[black] (0,\stepoffset-0.5) edge (gru1);
			\end{tikzpicture} 
		};
		
		\draw [arrow] (cluster1) -- node[anchor=south, yshift=0.1cm, xshift=-0.3cm] {\textit{generic}} (tree1.north);
		\draw [arrow] (cluster2) -- node[anchor=north, yshift=-0.2cm, xshift=-0.1cm] {\textit{generic}} (tree2.north);
		\draw [arrow] (cluster1) -- node[anchor=south, yshift=0.2cm, xshift=-0.1cm] {\textit{semantic}} (tree1.south);
		\draw [arrow] (cluster2) -- node[anchor=north, yshift=-0.1cm, xshift=-0.4cm] {\textit{semantic}} (tree2.south);
		
		\node[rectangle, text centered, text width=2cm, right of=tree1, fill=white, xshift=0.0cm](train) {\textbf{Training}};
		\node[rectangle, text centered, text width=2cm, right of=tree2, fill=white, xshift=0.0cm](val) {\textbf{Validation}};
		
		\end{tikzpicture}
	
		\captionsetup{width=.9\linewidth}
		{\caption{Overview of the model pipeline.\\
			First, we split our data into a training set (e.g., $90\%$) and a validation set (e.g., $10\%$).
			Then, we apply data augmentation on the training set to make it more robust (c.f.~section \ref{subsec.augmentation}).
			Then, we transform the formulae into trees (c.f.~section \ref{subsec.formulaparsing}).
			Then, we cluster the formula trees (c.f.~section \ref{subsec.clustering}).
			Finally, we train our neural network on the training data set (c.f.~section \ref{sec.training}) and evaluate it on the validation data set (c.f.~section \ref{sec.results}).
		}
		\label{fig:overview}}
	\end{figure}

	\subsection{Approach}

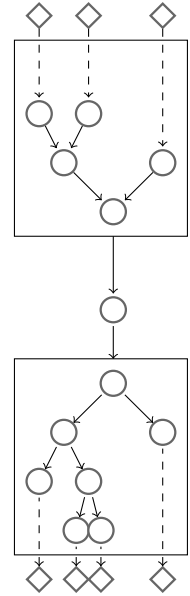
\begin{wrapfigure}{R}{0.3\textwidth}
  		\centering

  		\usetikzlibrary{arrows}
  		\def\layersep{1}
  		\def\stepoffset{3.5}

  		\begin{tikzpicture}[shorten >= 1pt, ->, node distance=\layersep,scale=0.65]
  		\tikzstyle{neuron} = [circle, draw=black!60, line width=0.3mm, fill=white];
  		\tikzstyle{treenode} = [circle, draw=black!60, line width=0.3mm, fill=white];
  		\tikzstyle{particle} = [diamond, draw=black!60, line width=0.3mm, fill=white,scale=0.7];

  		\node[treenode] (tree1-h1) at (0,0) {};
  		\node[treenode] (tree1-h2) at (-1,-1) {};
  		\node[treenode] (tree1-h3) at (1,-1) {};
  		\path[black,<-] (tree1-h2) edge (tree1-h1);
  		\path[black,<-] (tree1-h3) edge (tree1-h1);
  		\node[treenode] (tree1-h4) at (-1.5,-2) {};
  		\node[treenode] (tree1-h5) at (-0.5,-2) {};
  		\path[black,<-] (tree1-h4) edge (tree1-h2);
  		\path[black,<-] (tree1-h5) edge (tree1-h2);
  		\node[treenode] (tree1-h6) at (-0.75,-3) {};
  		\node[treenode] (tree1-h7) at (-0.25,-3) {};
  		\path[black,<-] (tree1-h6) edge (tree1-h5);
  		\path[black,<-] (tree1-h7) edge (tree1-h5);
  		\draw (-2,-3.5) rectangle (1.5, 0.5);

  		\node[particle] (t1-v1) at (-1.5,-4) {};
  		\path[black,dashed,<-] (t1-v1) edge (tree1-h4);
  		\node[particle] (t1-v2) at (-0.75,-4) {};
  		\path[black,dashed,<-] (t1-v2) edge (tree1-h6);
  		\node[particle] (t1-v3) at (-0.25,-4) {};
  		\path[black,dashed,<-] (t1-v3) edge (tree1-h7);
  		\node[particle] (t1-v4) at (1,-4) {};
  		\path[black,dashed,<-] (t1-v4) edge (tree1-h3);

  		\node[neuron] (gru1) at (0,1.5) {};
  		\path[black,<-] (0,0.5) edge (gru1);
  		\node[treenode] (tree2-h1) at (0,\stepoffset-0) {};
  		\node[treenode] (tree2-h2) at (-1,\stepoffset--1) {};
  		\node[treenode] (tree2-h3) at (1,\stepoffset--1) {};
  		\path[black] (tree2-h2) edge (tree2-h1);
  		\path[black] (tree2-h3) edge (tree2-h1);
  		\node[treenode] (tree2-h4) at (-1.5,\stepoffset--2) {};
  		\node[treenode] (tree2-h5) at (-0.5,\stepoffset--2) {};
  		\path[black] (tree2-h4) edge (tree2-h2);
  		\path[black] (tree2-h5) edge (tree2-h2);
  		\draw (-2,\stepoffset+3.5) rectangle (1.5, \stepoffset-0.5);

  		\node[particle] (t2-v1) at (2.5-4,\stepoffset+4) {};
  		\path[black,dashed] (t2-v1) edge (tree2-h4);
  		\node[particle] (t2-v2) at (3.5-4,\stepoffset+4) {};
  		\path[black,dashed] (t2-v2) edge (tree2-h5);
  		\node[particle] (t2-v3) at (5-4,\stepoffset+4) {};
  		\path[black,dashed] (t2-v3) edge (tree2-h3);

  		\path[black] (0,\stepoffset-0.5) edge (gru1);

  		\end{tikzpicture}
  		\caption{Scheme of the tree-to-tree neural network (figure based on a figure from \cite{Louppe:2017ipp})}
  		\label{fig:event_embedding}
  	\end{wrapfigure}

	Figure \ref{fig:overview} gives an overview of our model pipeline.
	In the following, we will describe our recursive neural network approach as shown in Figure \ref{fig:event_embedding}.
	Here, the recursive topology of these LSTM networks depends on the structure of the input tree.
	In the following, we will use terminology from the ``Deep Learning'' book by Goodfellow et al.~\cite{Goodfellow-et-al-2016}.
	
	The input of our neural network is a batch of trees with the same topology.
	The ground truth of our neural network is a batch of trees with the same topology.
	
	A one-hot encoding represents the values of the nodes of the trees.
	The encoder (upper half of Figure \ref{fig:event_embedding}) is responsible for converting the binary tree into a hidden state, while the decoder (lower half of Figure \ref{fig:event_embedding}) is responsible for converting a hidden state into a new binary tree.

	For the encoder, we recursively obtain a hidden state by feeding the hidden states of the left and right child as well as the value of the node to one two-variate (aka.~two-ary) LSTM.
	The hidden states of the leaves are initialized as zero.
	Thus, we obtain a hidden state.
	
	We may apply a quasi-linear function to this hidden state;
	whether this improves the translation is not yet determined.
	This is mandatory if the hidden state sizes of the encoder and the decoder differ.
	
	For the decoder, we use one two-valued LSTM to generate the hidden state of the left and right child and predict the value of the node.
	Further, the decoder is fed with the prediction of its parent's value.
	This two-valued LSTM is equivalent to two separate LSTMs, one LSTM being responsible for generating the left output and the other for the right output.

	Since we apply an additional tree traversal on the binary trees for the purpose of obtaining a better clustering, we need to consider this traversal inside the neural network.
	Each time we take or obtain two children, we lookup the stored traversal information $\mathcal{T}$ where $\mathcal{T} = 0$ if the children were not exchanged and $\mathcal{T} = 1$ if the children were exchanged in the preprocessing step.
	Then, we compute
	$$ (1-\mathcal{T}) \cdot \left( U_{L} \textbf{h}_{k_L} + U_{R} \textbf{h}_{k_R} \right) + \mathcal{T} \cdot \left( U_{R} \textbf{h}_{k_L} + U_{L} \textbf{h}_{k_R} \right) $$
	instead of
	$$ U_{L} \textbf{h}_{k_L} + U_{R} \textbf{h}_{k_R} $$ 
	for the left and right children's weight matrices $U_{L}, U_{R}$ and the left and right children's hidden states $\textbf{h}_{k_L}, \textbf{h}_{k_R}$.
	By this replacement, the neural network obtains the correct trees, while the topology of the neural network is independent of a clustering improving traversal.
		
	We use this structure per layer, while the hidden state of a previous layer is the input of the next layer.
	The input of the first layer of the encoder is the embedding of the respective value in the input tree.
	The output of the highest layer is used for the prediction.
	The prediction is computed by the application of two embeddings on the output of a respective LSTM.
	We initialize the weight matrices using the Xavier initializer \cite{pmlr-v9-glorot10a}, we initialize the biases to zero.
	
	\subsubsection{Training}\label{sec.training}
	
	In this section, we describe how we trained the aforementioned recursive neural network.
	We use the clusters of trees as mini-batches.
	As loss function, we use softmax cross-entropy.
	
	Since the actual formula is only a subtree of the input for the neural networks with sizes of roughly about $30-60\%$, most nodes are padded with zeros and $\textbf{y}_\mathrm{end}$s.
	This causes the primary influence on the loss to be the prediction of the padding zeros and $\textbf{y}_\mathrm{end}$s.
	Thus, the neural network learns the local minimum of only predicting these recurrent values quickly.
	
	Furthermore, the high occurrence frequency of tokens such as,~``\verb@^@'' in mathematical formulae yields a strong bias towards predicting these tokens.
	
	Therefore, we mask the loss function to lower the influence of commonly occurring symbols, such as parentheses, on the learning process.
	We achieve this by element-wise multiplication of the losses with the mask $$ m_i := \begin{cases}
	\frac\beta{|| \{\textbf{y}_j | \textbf{y}_i = \textbf{y}_j \} ||^\alpha} & \text{if $\hat{\textbf{y}}_i = \textbf{y}_i$} \\
	\frac1{|| \{\textbf{y}_j | \textbf{y}_i = \textbf{y}_j \} ||^\alpha} & \text{if $\hat{\textbf{y}}_i \neq \textbf{y}_i$}
	\end{cases} $$ where $\hat{\textbf{y}}$ is the prediction and $\textbf{y}$ is the correct translation.
	Here, $\alpha, \beta \in \left[0, 1\right]$.
	$\alpha$ sets the strength of the masking. While $\alpha=0$ causes the mask to have no influence, $\alpha=1$ causes the influence of a loss to be the reciprocal of the number of occurrences of its symbol in the mini-batch.
	$\beta$ is responsible for a reduction of the influence iff the prediction is correct.
	A too low $\beta$ may cause oscillation.
	
	Because the speed of learning should be independent of the size of the input tree's size and the local mini-batch size, we use \verb@reduce_sum@ instead of \verb@reduce_mean@ in order to weight every node equally and independent of the mini-batch in which it is included.
	The advantages of the common \verb@reduce_mean@, easier search for a learning rate and better numeric stability, are outweighed in this case.
	
	For training, we use the RMSProp optimizer \cite{DBLP:journals/corr/Ruder16} and the Adam optimizer \cite{KingmaB14} with learning rates between $10^{-6}$ and $10^{-4}$.
	The learning rate for the Adam optimizer has to be lower than for the RMSProp optimizer.
	As the optimal dropout, we found that $ 5 \% $ performs best.
	
	As an alternative to directly feeding the encoder with the input formula and the decoder with the output formula, we can use two auto-encoders, one for the input formula and another one for the output formula.
	This means that the encoder, as well as the decoder, is fed with the same values.
	Thus, the auto-encoders are trained to simulate the identity function.
	Because the information has to be compressed to be preserved while being propagated through the neural network as a state of lower size, a fitting representation is learned.
	Subsequently, we can combine the auto-encoder of the input formula as the encoder and the auto-encoder of the output formula as the decoder to a new neural network.
	By having translators from an input tree to a vector and from another vector to an output tree,	we can easily train a mapping between these representations.
	
	\subsubsection{Standalone disambiguation}
	Since mathematical notations, as well as an element-wise mapping from \textit{generic} to \textit{semantic} \LaTeX{}, are ambiguous, we implemented a standalone disambiguation.
	This disambiguation decides - for a given \textit{generic} \LaTeX{} symbol and a bag-of-words of the formula - which \textit{semantic} \LaTeX{} macro is correct.
	
	We restrict the classification to predictions that make sense.
	I.e., we choose only between those \textit{semantic} \LaTeX{} macros that would be re-translated into the same \textit{generic} \LaTeX{} symbol.
	
	We trained several neural networks with one to five hidden layers on this classification task.
	
	\subsubsection{Implementation}
	
	At firsts, we implemented a single-layer version of the neural network in TensorFlow.
	We found that in our case plain Python resulted in a speed increase of about 10 to 50 times compared to TensorFlow.
	TensorFlow was so slow because the recurrent changing of the neural network topology takes a long time in TensorFlow.
	Thus, a majority of the training time has been used by the creation of the neural network because we needed to recreate the neural network for each mini-batch.
	Further, TensorFlow is optimized for GPUs, yet we used CPUs because of a high memory demand.
	Therefore, we rewrote our TensorFlow code in plain Python utilizing numpy and autograd.
	While the TensorFlow implementation consumed up to $ 500 $GB of RAM for a single layer, our numpy implementation uses only up to $ 12 $GB of RAM with three layers.

	\section{Data and Preprocessing}\label{sec.data}
	
	This section describes the preprocessing steps from formula strings to clusters of formula trees.
	
	For training and evaluation, we used $13\,939$ \textit{generic} / \textit{semantic} \LaTeX{} formula pairs from the DLMF/DRMF, NIST.
	As shown in Figure \ref{fig:overview}, we first apply data augmentation on the training set.
	Then, we parse the formulae to trees and cluster these trees with respect to their topology.
	
	\subsection{Data augmentation}\label{subsec.augmentation}
	To achieve a robust neural network, we extended our dataset using the following mechanism.
	For this, we split our dataset on comparators\footnote{We use \verb@=, \gt, \lt, \equiv, \asympeq, \sim, \[Rr]ightarrow,@ and \verb@\to@} since the subexpressions obtained by these splits are still valid mathematical expressions.
	We then take the single terms, every two and three adjacent terms, including one or two comparators, and the whole formula (less if less is given).
	The following example shows these steps.
	As a result of this data augmentation, we received a data set containing $51\,217$ mathematical terms.
	
	\begin{alignat*}{6}
		& -\tfrac{3}{2}{\pi}+\delta\ & \leq\ & \mathrm{ph} { ((\lambda_{2}-\lambda_{1})z)}\ & \leq\ & \tfrac{3}{2}{\pi}-\delta \\
		\rightarrow{}\quad & & & & & \\
		1.\quad & -\tfrac{3}{2}{\pi}+\delta & & & & \\
		2.\quad & & & \mathrm{ph} { ((\lambda_{2}-\lambda_{1})z)} & & \\
		3.\quad & & & & & \tfrac{3}{2}{\pi}-\delta \\
		4.\quad & -\tfrac{3}{2}{\pi}+\delta & \leq\ & \mathrm{ph} { ((\lambda_{2}-\lambda_{1})z)} & & \\
		5.\quad & & & \mathrm{ph} { ((\lambda_{2}-\lambda_{1})z)} & \leq\ & \tfrac{3}{2}{\pi}-\delta
	\end{alignat*}
	
	\subsection{Formula Parsing and Tree Traversal}\label{subsec.formulaparsing}
	
	In the following, we will describe our heuristics based formula parser for transforming the formulae into only-leaf-labeled binary parse trees.
	
	Our parser first tokenizes the formula string.
	It then recursively transforms the token stream into a non-full n-ary tree.
	Structure-only generating braces and brackets are omitted, since the structure of the tree represents them.
	The following heuristics can either be activated or deactivated; due to lack of time, we have not yet thoroughly evaluated, how well these heuristics perform.
	In future work, we will carry out this evaluation.
    To achieve the results reported on in this paper, we have activated the first two heuristics.
	
	\begin{itemize}
		\item If a node forms a command, \verb@<COMMAND_END>@ is added as an additional terminating leaf.
		\item If a node forms a concatenation on the same level (e.g., \verb@a + b@), \verb@<CONCAT_END>@ is added as an additional terminating leaf. 
		\item Leaves that are infix operators are moved to the left, such that they are present in prefix notation.
	\end{itemize}

	After applying this tree traversal, we transform the n-ary tree into an adaptation of left-child right-sibling binary trees.
	I.e., we adapt left-child right-sibling binary trees to leaf-only valued trees.
	Here, the left child is the first child, and the right child is the same node excluding the first child.

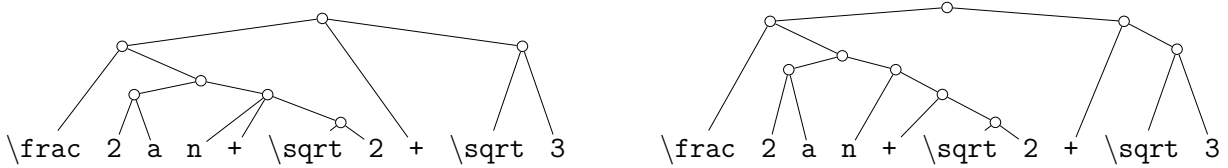
\begin{figure}[h!]
	\centering
	\forestset{
		style1/.style={
			for tree={
				inner sep=0.15em,
				l=0.0em,
				l sep=0.1em,
				if n children=0{
					tier=word
				}{
					circle, draw
				}
			}
		}
	}
	\noindent\resizebox{0.95\linewidth}{!}{
		\begin{forest}
			[, style1, [ [\texttt{\large $\backslash$frac}] [ [ [\texttt{\large 2}] [\texttt{\large a}] ] [ [\texttt{\large n}] [\texttt{\large +}] [ [\texttt{\large $\backslash$sqrt}] [\texttt{\large 2}] ] ] ] ] [\texttt{\large +}] [ [\texttt{\large $\backslash$sqrt}] [\texttt{\large 3}] ] ]
		\end{forest}
	\hspace{1cm}
		\begin{forest}
			[, style1, [ [\texttt{\large $\backslash$frac}] [ [ [\texttt{\large 2}] [\texttt{\large a}] ] [ [\texttt{\large n}] [ [\texttt{\large +}] [ [\texttt{\large $\backslash$sqrt}] [\texttt{\large 2}] ] ] ] ] ] [ [\texttt{\large +}] [ [\texttt{\large $\backslash$sqrt}] [\texttt{\large 3}] ] ] ]
		\end{forest}
	}
	\caption{Parse tree without \texttt{<COMMAND\_END>}, \texttt{<CONCAT\_END>} or infix to prefix traversal (left is n-ary, right is binary)}
	\label{figure:tree-without}
	\end{figure}

\begin{figure}[h!]
		\centering
		\forestset{
			style1/.style={
				for tree={
					inner sep=0.15em,
					l=0.0em,
					l sep=1.0em,
					if n children=0{
						tier=word
					}{
						circle, draw
					}
				}
			}
		}
		\noindent\resizebox{1\linewidth}{!}{
			\begin{forest}
				[, style1, [ [,inner sep=0,l=1em[\texttt{\large $\backslash$frac}]] [ [ [\texttt{\large 2}] [\texttt{\large a}] [\texttt{\large <2>}] ] [ [\texttt{\large n}] [\texttt{\large +}] [ [\texttt{\large $\backslash$sqrt}] [\texttt{\large 2}] [\texttt{\large <1>}] ] [\texttt{\large <2>}] ] ] [\texttt{\large <1>}] ] [\texttt{\large +}] [ [\texttt{\large $\backslash$sqrt}] [\texttt{\large 3}] [\texttt{\large <1>}] ] ]
			\end{forest}
\hspace{1cm}
		\forestset{
			style1/.style={
				for tree={
					inner sep=0.15em,
					l=0.0em,
					l sep=0.1em,
					if n children=0{
						tier=word
					}{
						circle, draw
					}
				}
			}
		}
			\begin{forest}
				[, style1, [ [,inner sep=0,l=1em[\texttt{\large $\backslash$frac}]] [ [ [ [\texttt{\large 2}] [ [\texttt{\large a}] [\texttt{\large <2>}] ] ] [ [\texttt{\large n}] [ [\texttt{\large +}] [ [ [\texttt{\large $\backslash$sqrt}] [ [\texttt{\large 2}] [\texttt{\large <1>}] ] ] [\texttt{\large <2>}] ] ] ] ] [\texttt{\large <1>}] ] ] [ [\texttt{\large +}] [ [\texttt{\large $\backslash$sqrt}] [ [\texttt{\large 3}] [\texttt{\large <1>}] ] ] ] ]
			\end{forest}
		}
		\caption{Parse tree with \texttt{<COMMAND\_END>} as \texttt{<1>} and \texttt{<CONCAT\_END>} as \texttt{<2>} (left is n-ary, right is binary)}
		\label{figure:tree-end-binary}
	\end{figure}
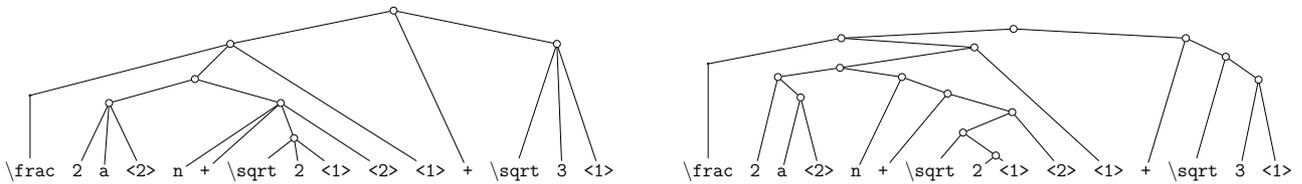

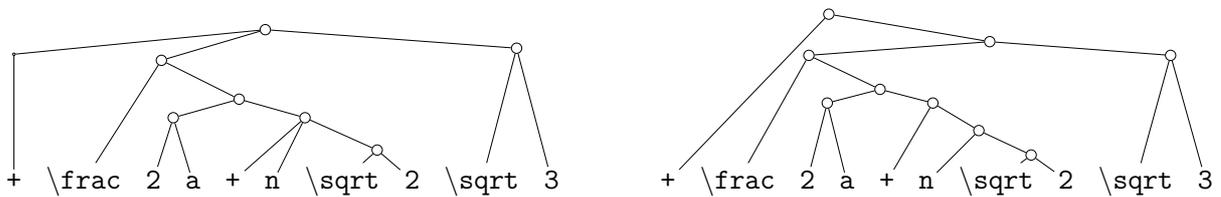
\begin{figure}[h!]
	\centering
	\noindent\resizebox{0.95\linewidth}{!}{
		\forestset{
			style1/.style={
				for tree={
					inner sep=0.15em,
					l=0.0em,
					l sep=0.3em,
					if n children=0{
						tier=word
					}{
						circle, draw
					}
				}
			}
		}
		\begin{forest}
			[, style1, [,inner sep=0,l=0.25em[\texttt{\large +}]] [,l=0.5em [\texttt{\large $\backslash$frac}] [ [ [\texttt{\large 2}] [\texttt{\large a}] ] [ [\texttt{\large +}] [\texttt{\large n}] [ [\texttt{\large $\backslash$sqrt}] [\texttt{\large 2}] ] ] ] ] [ [\texttt{\large $\backslash$sqrt}] [\texttt{\large 3}] ] ]
		\end{forest}
	\hspace{1cm}
		\forestset{
			style1/.style={
				for tree={
					inner sep=0.15em,
					l=0.0em,
					l sep=0.1em,
					if n children=0{
						tier=word
					}{
						circle, draw
					}
				}
			}
		}
		\begin{forest}
			[, style1, [\texttt{\large +}] [ [ [\texttt{\large $\backslash$frac}] [ [ [\texttt{\large 2}] [\texttt{\large a}] ] [ [\texttt{\large +}] [ [\texttt{\large n}] [ [\texttt{\large $\backslash$sqrt}] [\texttt{\large 2}] ] ] ] ] ] [ [\texttt{\large $\backslash$sqrt}] [\texttt{\large 3}] ] ] ]
		\end{forest}
	}
	\caption{Parse tree with infix to prefix traversal (left is n-ary, right is binary)}
	\label{figure:tree-infix}
	\end{figure}

	The following example input illustrates the parsing and traversal process:
	\begin{center}
		\verb@\frac{2a}{n+\sqrt{2}}+\sqrt{3}@ $\quad \left( \frac{2a}{n+\sqrt{2}}+\sqrt{3} \right)$
	\end{center}
	First, we tokenize the input:
	\begin{center}
		\verb@\frac { 2 a } { n + \sqrt { 2 } } + \sqrt { 3 }@
	\end{center}
	As shown in Figure \ref{figure:tree-without}, we generate an n-ary tree from this token stream.
	Next, we can (optionally) add the \verb@<COMMAND_END>@ and \verb@<CONCAT_END>@ tokens as shown in Figure \ref{figure:tree-end-binary}.
	Additionally, we can (optionally) traverse the tree from infix to prefix notation as shown in Figure \ref{figure:tree-infix}.

	Now, we can perform a traversal to gain the right-child-biggest property.
	For this, we recursively exchange the children of a tree if the size of the left child is greater than that of the right child.
	For a reconstruction within the neural network, we store the information on which children of a tree we performed the exchange.
	Thus, we obtain trees that are more similar, which improves the clustering (i.e., increases the speed of training).
	
	After the subsequent clustering, these binary trees serve as input for the neural network.

	\subsection{Tree clustering}\label{subsec.clustering}
	
	Because the topology of the neural network depends on the input, we would have to generate a new neural network for each formula.
	Therefore, we apply a padding to the binary trees.
	To do this, we create a binary tree $a$ with a topology greater or equal to every tree $\mathbf{x} \in \mathbf{X}$ and compute $\textbf{x}_a$, the padded version of $\textbf{x}$. 
	
	Generating the neural network for $a$ for all trees, would result in having $\sim 80\,000$ nodes which would require approximately tens of Terabytes of RAM, which is impractical.
	Therefore, we cluster the pairs of binary trees $(\mathbf{x}, \mathbf{y})$ according to their topology.
	
	We implemented numerous standard clustering algorithms, including \verb@k-means@, \verb@DBSCAN@, and \verb@linkage@-algorithms.
	The results of these algorithms were insufficient.
	Therefore, we implemented Algorithm \ref{algorithm:pick_clustering}.
	For this algorithm, the choice of \text{min-elems-list} and \text{max-size-list} is crucial.
	The following choices yielded the best results:\\
	\begin{align}
	(\text{min-elems})_i &:= \left\lfloor \frac{2\% \cdot \#_\text{trees to cluster}}{\sqrt[||\text{min-elems}||-1]{2\% \cdot \#_\text{trees to cluster}}^{i-1}} \right\rfloor \label{eq:min-elems} \\
	(\text{max-size})_i &:= \left\lfloor 5 \cdot \left( \sqrt[||\text{max-size}||-1]{10^4} \right)^{i-1} \right\rfloor \label{eq:max-size}
	\end{align}
	
	\begin{algorithm}
		\SetAlgoLined
		\KwData{masks}
		\KwResult{clustering}
		clustering $\gets$ \text{empty clustering}$\;$\\
		min-elems-list $\gets$ \text{descending list} \text{ (e.g.,~20 exponential steps $\subset [\lfloor 2\% \cdot \#_\text{trees to cluster} \rfloor, 1]$)}$\;$\\
		max-size-list $\gets$ \text{ascending list} \text{ (e.g.,~20 exponential steps $\subset [5, 5\cdot10^4]$)}$\;$\\
		\For{ min-elems, max-size \textbf{in} min-elems-list, max-size-list}{
			clustering $\gets$ \textbf{FindClusters} \text{(clustering, masks, min\-elems, max\-size)}$\;$\\
		}
		\caption{Pick clustering}
		\label{algorithm:pick_clustering}
	\end{algorithm}
	\begin{algorithm}
		\SetAlgoLined
		\KwData{clustering, masks, min-elems, max-size}
		\KwResult{clustering}
		\For{mask \textbf{in} non-clustered masks}{
			initialize new cluster$\;$\\
			\While{$\min \{\ \|a\|\  |\ \forall e \in \text{cluster } : e \subset a \}$ $\leq$ max-size}{
				Add all elements to cluster that do not increase $\|a\|$$\;$\\
				Add the element to cluster which has the least increment to $\|a\|$ if $\min \{\ \|a\|\  |\ \forall e \in \text{cluster } : e \subset a \}$ $\leq$ \textit{max-size} still holds$\;$\\
			}
			\If{number of elements in cluster $<$ min-elems}{
				resolve cluster$\;$
			}
		}
		\caption{Find clusters}
	\end{algorithm}

	Compared to the other tested clustering algorithms, this algorithm led to the best results on the given task of finding clusters with a small hull tree and finding as many elements as possible per cluster in a short time.
	The table in Figure \ref{table:Comparison-clustering-algorithms} shows selected results of the clustering algorithms.
	Experiments showed that the training time is roughly proportional to
	\begin{equation} \label{eq:training time}
		\#\text{clusters} \cdot \left( \textstyle \sum_{\textbf{x}\in \textbf{X}} ||\ a_\text{\textbf{x}`s cluster}\ ||\right)^\alpha
		\qquad \text{where }\alpha \in [1,1.02]
	\end{equation}
	and where $a_\text{\textbf{x}`s cluster}$ is the minimal tree, such that $a\supset b$ holds for each $b$ in \textbf{x}'s cluster.
	Thus, we want to minimize this product to achieve a shorter training duration.
	In general, it is more important to reduce the number of clusters than the size of the paddings because this enables a better parallelization and because less native Python code is executed.
	By combining this clustering algorithm and the previously presented tree traversal, semi-batched computation and training is made possible.
	
	\begin{figure*}
	\centering
	\begin{minipage}{1.0\linewidth}
		\resizebox{1.0\linewidth}{!}{
		\begin{tabular}{|l l c c c c|} 
			\hline &&&&&\\ [-2ex]
			Algorithm & parameters & $\#\text{clusters}\ $\footnote{Less important quality measure: Lower is better}\  & $\sum_{x\in X} ||\ a_\text{$x$`s cluster} \ ||\ $\footnote{More important quality measure: Lower is better}\  & est. training time\footnote{According to equation \ref{eq:training time} with $\alpha=1$ and multiplied with $ 10^{-6} $} & CPU-time\footnote{CPU-time of clustering including preprocessing and evaluation on SCCKN at $\sim 2.5\text{GHz}$} \\ [0.5ex] 
			\hline\hline 
			&&&&&\\ [-2ex]
			no clustering &  & $15\,240$ & $1 \,839\,066 $ & $ 28\,027 $ & - \\
			Algorithm \ref{algorithm:pick_clustering} & \texttt{parameters 1}\footnote{As in equations \ref{eq:min-elems} and \ref{eq:max-size}} & $541$ & $ 4\,913\,814 $ & $ 2\,658 $ & 02:41:09 \\
			Algorithm \ref{algorithm:pick_clustering} & \texttt{parameters 2} & $1\,815$ & $ 2\,998\,306 $ & $ 5\,441 $ & 01:27:49 \\
			Algorithm \ref{algorithm:pick_clustering} & \texttt{parameters 2}, no tree traversal & $2\,071$ & $ 3\,298\,798 $ & $ 6\,832 $ & 08:48:44 \\
			\texttt{k-means} & $k=935$ & $ 935 $ & $ > 5\,222\,216 $ & $ > 5\,000 $ & 02:09:45 \\
			\texttt{k-means} & $k=2\,287$ & $2\,287$ & $ > 2\,817\,670 $ & $ > 6\,500 $ & 123:10:20 \\
			\texttt{complete-linkage} & $n=2\,287$ & $ 2\,287 $ & $ > 4\,187\,086 $ & $ > 9\,575 $ & 01:04:07 \\
			\texttt{DBSCAN}\footnote{\texttt{DBSCAN} was very weak because either a lot of elements were not related to a cluster or the clusters were too huge.} & various options & $ >7\,000 $ & $ >10\,000\,000 $ & $ > 70\,000 $ & - \\
			 [1ex] 
			\hline
		\end{tabular}
		}
	\caption{
		Comparison of clustering algorithms.
		The application of tree traversal is implied.
		For performance reasons, the clusterings are only computed for $15\,240$ of the $51\,217$ mathematical terms in our data set.
	}
	\label{table:Comparison-clustering-algorithms}
	\end{minipage}
	\end{figure*}

	\section{Results}\label{sec.results}
	
	We evaluated our recursive neural network using 10-fold cross validation.
	As the accuracy measure, we computed full accuracy $p_\mathrm{f}$ and masked accuracy $p_\mathrm{m}$.
	\begin{align}
		p_\mathrm{f} &:= \frac{|| \{\ i\ |\ \hat{\textbf{y}}_i = \textbf{y}_i\ \} ||}{|| \{\ i\ |\ \textbf{y}_i \in \textbf{y}\ \} ||} 
		& p_\mathrm{m} &:= \frac{|| \{\ i\ |\ \hat{\textbf{y}}_i = \textbf{y}_i, \textbf{y}_i \neq 0, \textbf{y}_i \neq \textbf{y}_\mathrm{end}\ \} ||}{|| \{\ i\ |\ \textbf{y}_i \neq 0, \textbf{y}_i \neq \textbf{y}_\mathrm{end}\ \} ||}
	\end{align}
	The accuracies ignore that a prediction may be almost correct if the result is, for example, shifted by one node.
	Therefore, if an error occurs near the root, a proper subtree at the wrong position is interpreted as wrong, too.
	For this reason, we additionally implemented a bag-of-words accuracy
	\begin{equation}
		p_\mathrm{b} := 1 - \sum_{i=0}^{n-1} \left( \frac{\max\left(\#_i^{\textbf{y}}-\#_i^{\hat{\textbf{y}}}, 0 \right)}{\| \textbf{y} \|} \right)
	\end{equation}
	where $n$ is the number of possible values, and $\#_i^{\textbf{y}}$ is the number of occurrences of $i$ in $\textbf{y}$.
	The bag-of-words accuracy thus measures whether the prediction consists of the same unsorted values and frequencies as the ground truth but ignores their structure.
	This makes the measure robust against shifts of nodes.
	Since the training is in no way biased towards the direction of this measure, it is very likely that most of the errors which cause a reduction in the full accuracy, but are correct with respect to the bag-of-words accuracy, are only a slight displacement within the tree.
	
	If the full accuracy is higher than the masked accuracy, the prediction of the topology of the tree (equal to the prediction of zeros) is better than the prediction of the values.
	This can indicate a bias towards predicting zeros.
	While this almost always occurs, our loss function avoids falling into a local minimum.
	
	Until now, we mainly trained on a subset of only $15\,240$ mathematical terms for performance reasons.
	Further, we used relatively high learning rates.

	Currently, we achieved a masked validation accuracy of $p_\mathrm{m} = 47.05\%$.
	Further, we repeatedly obtained full validation accuracies of up to $p_\mathrm{f} = 80\%$.
	Our experiment with the best masked validation reached a bag-of-words accuracy of $p_\mathrm{b} = 92.3\%$.

	The results show that the single-layer approach outperforms the multi-layer approaches.
	An LSTM state size of 256 produced the best results.\\

	By traversing and clustering the trees, we could achieve a speed increase of roughly factor ten.
	
	\subsection{Standalone disambiguation}
	
	The standalone disambiguation of mathematical formulae, which chooses which mathematical interpretation (i.e.,~\textit{semantic} \LaTeX{} macro) to use, yielded excellent results.
	Without training (i.e.,~random prediction), we achieved an accuracy of $50-60\%$.
	This is because some \textit{generic} \LaTeX{} symbols are restricted to a single \textit{semantic} \LaTeX{} macro.
	Using our bag-of-words approach and neural networks, we obtained an accuracy of about $97\%$ per choice of a \textit{semantic} \LaTeX{} macro.
	
	Since we would need a rule-based translator which only utilizes the neural network in ambiguous cases, our standalone disambiguation approach is currently not applicable.
	Our results show that neural networks are adequate for this disambiguation task.
	Thus, if a rule-based translator with ambiguities will be implemented, this would be a good solution.

	\section{Discussion and Conclusion} \label{sec.disconcl}

	The current results cannot yet be applied to a complete translation but are promising considering that we did not yet use an attention mechanism, and that we did not yet use the entire data set for training.
	Further, a shortage of time did not allow a full training.

	One major drawback of recursive neural networks is their performance, because, batched training is generally not possible.
	We achieved a speed increase of roughly factor ten.
	How well our technique can be combined with other speed-optimizing approaches remains to be determined.

	A drawback of clustering the training data is that the order of occurring formulae is biased, because formulae with a similar topology are used together for training.
	Experiments with a different neural network on the same bias showed that this bias can be neglected if the learning rate is low enough.
	We think that the single-layer network outperforms the multi-layer network since pre-trained embeddings are not available, and since we didn't perform layer-wise training, yet.

	\section{Outlook}\label{sec.outlook}
	
	In future research, we will improve our implementation to enable a complete formula translation.
	For that, we will implement a novel attention mechanism for recursive neural networks.
	Additionally, we will extend our auto-encoder approaches and will apply layer-wise training.

	For the parsing and tree traversal, we will implement an improved traversal approach based on post-fix notation to enhance the tree structure.

	In the future, this work may allow the DLMF/DRMF to extend their repository of \textit{semantic} \LaTeX{} formulae more easily.
	Other applied use cases for this translator are mathematical spell-checkers, validation of mathematical books, easier interfaces for CASs, and the automated pre-correction of students' math problem sets.\\
	
	The standalone disambiguation could also be implemented using a recurrent or recursive neural network instead of a bag-of-words based approach.
	This would most likely yield better results, because the order of the formulae, or respectively, the structure of the formulae can be considered using this approach.

	\section*{Acknowledgments}

		We would like to thank Howard Cohl et al.~\cite{Cohl15} for sharing their unreleased \textit{semantic} \LaTeX{} macros including $13\,939$ formulae from the DLMF/DRMF in this format.
		Further, we would like to thank Christian Borgelt for advising on the domain of neural networks.
		This work was supported by the FITWeltweit program of the German Academic Exchange Service (DAAD), as well as the
		{German Research Foundation (DFG}, grant %
		{GI-1259-1}).
		
%
	
	%
	
	%
	
%
%


	%
	%
	
	%
	%
	%
	%
	
\end{document}